\documentclass[runningheads]{llncs}

\usepackage{eccv}   

\usepackage{eccvabbrv}
\usepackage{placeins}
\usepackage{graphicx}
\usepackage{booktabs}
\usepackage{multirow}
\usepackage[accsupp]{axessibility}  % Improves PDF readability for those with disabilities.

\usepackage{hyperref}
\usepackage{orcidlink}
\usepackage{caption}
\usepackage[table]{xcolor}

\begin{document}

\title{TAU-R1: Visual Language Model for Traffic Anomaly Understanding}

\titlerunning{TAU-R1}

% ORCID: add \orcidlink{0000-0000-0000-0000} after each author name once collected.
\author{Yuqiang Lin\inst{1}\thanks{Corresponding authors.} \and
Kehua Chen\inst{2} \and
Sam Lockyer\inst{1} \and
Arjun Yadav\inst{3} \and
Mingxuan Sui\inst{1} \and
Shucheng Zhang\inst{2} \and
Yan Shi\inst{2} \and
Bingzhang Wang\inst{2} \and
Yuang Zhang\inst{2} \and
Markus Zarbock\inst{4} \and
Florian Stanek\inst{4} \and
Adrian Evans\inst{1} \and
Wenbin Li\inst{1} \and
Yinhai Wang\inst{2} \and
Nic Zhang\inst{1}$^{\star}$}
\authorrunning{Y.\ Lin et al.}

\institute{University of Bath, Bath BA2 7AY, United Kingdom\\
\and
University of Washington, Seattle, WA 98105, USA \and
City of Carmel, Carmel, IN, USA \and
Starwit Technologies GmbH, Wolfsburg, Germany \and
\email{yl4300@bath.ac.uk} }

\maketitle
\begin{abstract}
Traffic Anomaly Understanding (TAU) is important for traffic safety in Intelligent Transportation Systems. Recent vision-language models (VLMs) demonstrate strong capabilities in video understanding. However, progress on TAU remains limited due to the lack of benchmarks and task-specific methodologies. To address this limitation, we introduce \textbf{Roundabout-TAU}, constructed from real-world roundabout videos collected in collaboration with the City of Carmel, Indiana. The dataset contains 342 clips and is annotated with 2,000+ question--answer pairs covering multiple aspects. Building on this benchmark, we propose \textbf{TAU-R1}, a two-layer vision--language framework for TAU. The first layer is a lightweight anomaly classifier that performs coarse anomaly categorisation, while the second layer is a larger anomaly reasoner that generates detailed event summarisation. To improve task-specific reasoning, a two-stage training strategy is introduced consisting of decomposed-QA enhanced supervised fine-tuning followed by TAU-GRPO, a GRPO-based post-training method with TAU task-specific reward functions. Experimental results demonstrate that TAU-R1 achieves strong performance on both anomaly classification and reasoning tasks while maintaining deployment efficiency. The dataset and code are released at: \url{https://github.com/siri-rouser/TAU-R1}.
\end{abstract}

\section{Introduction}
Traffic safety remains a central concern in modern intelligent transportation systems. Detecting and understanding traffic anomalies can support faster incident response, reduce secondary harm, improve public safety, and enhance transportation efficiency \cite{zhang2011data,xu2021sutd}. However, most existing vision-based traffic anomaly research \cite{song2025real,bao2020ustring,yao2019unsupervised,fang2022ssc} focuses on detection or prediction, where the output is usually only an anomaly score or a coarse event label. While useful for screening, such outputs provide limited semantic understanding of what happened, why it happened, and which road users were involved.

Recent advances in vision--language models (VLMs) \cite{Qwen3-VL,achiam2023gpt,lin2024video,feng2025video,xiao2023florence} have enabled increasingly nuanced video understanding. By combining visual perception with language reasoning, VLMs achieve strong performance on general video understanding tasks \cite{cheng2024videollama2,Maaz2024VideoGPT+,guo2025deepseek}. However, applying general-purpose VLMs to roadside TAU presents challenges. While general video understanding often requires broad open-world knowledge, TAU requires fine-grained reasoning on specific scenes (e.g.\ vehicle behaviours and interactions) that determine whether an event is anomalous. This requires domain-specific knowledge of traffic rules, scene context, and the relationships underlying anomalous traffic events. Such capabilities require domain-specific datasets that accurately capture the characteristics of these real-world traffic scenes.

This leads to the second challenge, the lack of suitable data. Existing anomaly understanding studies \cite{xing2025echotraffic,huang2025vad,zhang2025holmes,chen2025aligning} mostly rely on open-source datasets, web-mined videos, or datasets later extended with text annotations. Although useful, these resources are often biased toward edited or visually obvious events and rarely focus on fixed roadside surveillance scenes. Consequently, existing datasets provide limited support for fine-grained understanding of real-world traffic anomalies in city-scale surveillance settings, especially for question answering and summary generation grounded in scene context, object interactions, and anomaly causes.

To address these limitations, \textbf{Roundabout-TAU} is introduced as a benchmark for traffic anomaly understanding, built from real-world roundabout surveillance videos collected in collaboration with the City of Carmel. Roundabouts represent a particularly challenging traffic environment due to dense vehicle interactions, subtle lane-use behaviour, and the coexistence of both subtle and extreme anomaly events. Minor behavioural anomalies---such as incorrect yielding, improper lane usage, or hesitation within circulating traffic---often occur alongside more obvious events, such as collisions or abrupt manoeuvres. This combination creates a complex setting that requires models to develop more fine-grained insights. This dataset contains 342 video clips captured from 28 roadside cameras across diverse viewpoints and traffic conditions, including both anomalous and normal events. To support language-based training and evaluation, the dataset is annotated with 2,064 multi-aspect question--answer pairs covering environment perception, object grounding, anomaly classification, event description, temporal localisation, and event reasoning. To the best of our knowledge, Roundabout-TAU is the first real-world roadside traffic anomaly benchmark with comprehensive QA-style annotations, and the first benchmark specifically dedicated to roundabout-centric TAU.

Building on this benchmark, we propose \textbf{TAU-R1}, a two-layer framework for efficient traffic anomaly understanding. The first layer is a lightweight anomaly classifier that predicts a coarse traffic anomaly label and serves as an efficient screening stage for deployment under constrained resources. The second layer is a larger anomaly reasoner that generates detailed event summaries for anomalous clips, enabling deeper interpretation of the anomaly event. This hierarchical design is motivated by deployment efficiency, enabling rapid filtering of normal traffic while reserving detailed reasoning for anomalous events.

To further improve task-specific reasoning, we introduce a two-stage training strategy. First, we use \emph{decomposed-QA enhanced supervised fine-tuning}, which decomposes TAU into several question--answer sub-tasks so that the model can learn the prior scene knowledge required for traffic anomaly understanding. Second, we propose \emph{TAU-GRPO}, a TAU-oriented extension of GRPO with task-specific reward functions for anomaly classification and summarisation. Our main contributions are summarised as follows:
\begin{itemize}
    \item We introduce \textbf{Roundabout-TAU}, a real-world benchmark for traffic anomaly understanding built from roadside roundabout surveillance videos, with QA-style annotations for fine-grained scene and event understanding.
    \item We propose \textbf{TAU-R1}, a two-layer framework that combines classification with deeper anomaly reasoning for practical traffic anomaly understanding.
    \item We develop a task-specific training strategy that combines \textbf{decomposed-QA enhanced supervised fine-tuning} with \textbf{TAU-GRPO}, a TAU-specific GRPO-based post-training method to further improve reasoning ability.
    \item Experiments show that the proposed framework achieves strong performance on both anomaly classification and anomaly summarisation, while the hierarchical design supports real-time edge deployment.
\end{itemize}
\section{Related Work}
\subsection{Video Anomaly Dataset}
Video anomaly datasets \cite{ucf-crime,xd-violence,ucsd-ped1,CUHK_Avenue,yao2022dota,shanghaitech,UBnormal,MSAD,DAD,TAD} were typically designed for anomaly detection, not understanding, and therefore provide coarse supervision such as binary labels or anomaly categories. These datasets are useful for detection benchmarking, but they do not support deeper understanding of what happened, why it happened, or which road users were involved. More recently, several works \cite{yuan2024towards,chen2023tevad,zhang2024holmes,lv2024video,bose2023mm,xing2025echotraffic} have started to move toward anomaly understanding by adding textual descriptions or QA-style annotations, enabling tasks like event description, reasoning, and question answering.

Within these datasets, multi-category benchmarks such as HIVAU-70k \cite{zhang2025holmes} which built upon previous open-source anomaly datasets such as UCF-Crime \cite{ucf-crime} and XD-Violence \cite{xd-violence}, include a small number of traffic-related incidents, but traffic is not their main focus and the number of traffic samples is limited. Only a few existing datasets are traffic-specific \cite{bose2023mm,xing2025echotraffic}; however, most of them are collected in dash-cam or ego-centric settings instead of fixed roadside surveillance views. Furthermore, many traffic anomaly datasets are constructed from web-mined videos (e.g., news, social media, and video-sharing platforms), which are often short, edited, and biased toward visually salient incidents. This bias limits coverage of subtle but safety-critical anomalies and makes it difficult to support fine-grained QA relating to causes, rule violations, and interactions. Consequently, QA-supervised roadside traffic anomaly resources remain limited in both scale and consistency, motivating the need for real-world fixed-camera benchmarks with comprehensive language annotations.
\subsection{Video Anomaly Understanding}
With the rapid development of Multi-Modal Large Language Model (MLLM), researchers have started to utilise their advanced reasoning ability to solve the VAU task. Among those methods, there are two main categories. The first category comprises prompt-based training-free methods \cite{yang2024follow, zanella2024harnessing,li2025vadtree,lin2025unified,ahn2025anyanomaly,shao2025eventvad,yang2025monitor} where a MLLM is used directly via prompts---for example as an anomaly scorer \cite{shao2025eventvad, zanella2024harnessing}, or a rule-based reasoning agent with few-shot examples \cite{yang2024follow}. These methods are simple to deploy, but often struggle when applied to traffic video, as the task requires very detailed domain-specific prior knowledge. The second category comprises training-based approaches \cite{zhang2025holmes, tang2024hawk,ye2025vera,zhanglvlms,chen2025aligning,ding2025slowfastvad,li2025lad,huang2025vad,xing2025echotraffic} that improve the anomaly understanding ability of general VLMs by using supervised fine-tuning or RL-based post-training. Among those training-based methods, some \cite{chen2025aligning,tang2024hawk,ding2025slowfastvad} improve visual grounding by extracting more representative visual tokens, while others \cite{zhang2025holmes} decompose anomalies into hierarchical levels or subtasks to make learning more structured. The only benchmark that is applied specifically to TAU tasks is EchoTraffic \cite{xing2025echotraffic}, which focuses mainly on dash-cam scenarios and emphasises audio-visual fusion rather than fixed roadside surveillance.

\section{Roundabout-TAU Dataset}
In this section, we introduce the Roundabout-TAU dataset by first defining the TAU task, and then describing the dataset construction and labelling process, followed by dataset statistics. To the best of our knowledge, Roundabout-TAU is the first traffic anomaly understanding dataset dedicated to roundabout scenes and the first TAU dataset collected from real-world traffic cameras.
\subsection{TAU Task Definition}
We define the traffic anomaly understanding task by considering the practical needs of a real-world traffic anomaly understanding system. In real deployments, city managers generally expect a TAU system to answer two key questions: \textit{(1) Is there an anomaly?} \textit{(2) If so, what happens?} We consider the first question as a classification problem that serves as the first step of TAU. The second question goes beyond anomaly detection. Given that an anomaly has already been detected, the goal is to understand the event in detail and answer what, why, and how questions about the anomaly.

\subsubsection{Classification} For the anomaly event classification task, following previous widely used benchmarks such as UCF-Crime \cite{ucf-crime} and XD-Violence \cite{xd-violence}, we pre-define four anomaly classes and ask the model to choose one of them. Those four classes are: \textit{(A) No anomaly}; \textit{(B) Direction or manoeuvre violation}, including wrong-way driving, illegal or unsafe turns, and illegal or unsafe lane changes; \textit{(C) Near-collision or collision event}, including cut-offs that force hard braking or evasive action, near-misses, and crashes; \textit{(D) Abnormal road use}, including stopping in an active lane or mid-road (breakdown, yielding to emergency vehicles, letting pedestrians cross) and driving or parking in non-vehicle areas (sidewalk, shoulder, etc.).

\subsubsection{Summarisation} For the anomaly event summarisation, instead of following previous benchmarks such as EchoTraffic \cite{xing2025echotraffic} and CUVA \cite{du2024uncovering}, which decompose video anomaly understanding into multiple questions including what, why, when, and how, we argue that a real-world traffic anomaly system only needs one comprehensive summary that clearly describes the anomaly event and explains its underlying factors. Therefore, we define our main TAU task as a single QA task, where the model is asked to provide a comprehensive event summary in a few sentences for anomaly events. The summary is designed to cover four perspectives: \textit{(1) Environment Perception}, including time of day, weather, road surface condition, and road topology; \textit{(2) Object Grounding}, including vehicle type, color, location in the scene (e.g.\ entry lane), and location in the frame (e.g.\ top-left) of the main anomalous objects; \textit{(3) Event description}, covering the main movements and interactions of the anomalous objects; and \textit{(4) Event analysis}, explaining the underlying reason behind the anomaly.

\subsection{Dataset Construction and Labelling}
\subsubsection{Video Source Collection}
We collect all traffic anomaly videos in collaboration with the City of Carmel, Indiana. The anomaly clips in our dataset come from two sources: (1) 147 anomaly clips from archived traffic incident videos collected during 2024--2026, and (2) 129 anomaly clips mined from an existing edge traffic anomaly detector deployed in real-world traffic scenes. The second set of clips was collected by the movement-predictor detector \cite{starwit_movement_predictor_2025}, which uses a traditional object-centric unsupervised method to identify video clips that deviate from normal traffic patterns. We retrieve 1,000 candidate videos from this process, manually inspect them, and finally select 129 high-quality clips. To improve generalisation and mitigate false positives, videos containing no anomalous events were also included. Normal clips were sampled for each site in proportion to the number of anomalous clips in the anomaly set, while ensuring that each site contributed at least two clips captured under different times of day and weather conditions. In total, 66 normal video clips were selected. Overall, our dataset contains 342 video clips from 28 different cameras, covering diverse resolutions and viewing angles in real-world traffic scenes.

\subsubsection{Question-Answer Pair Labelling}
For the data-labelling process, we develop a multi-stage annotation pipeline that combines human annotation with MLLM-assisted enrichment to ensure high-quality annotations. Instead of labelling only the anomaly class and a single event summary, we also provide decomposed annotations for environment, object grounding, event description, and event analysis, so that the dataset can support finer-grained traffic anomaly understanding.

In \textit{Stage 1}, we manually inspect every video, assign an anomaly category, and write a brief event summary. We choose to start from human annotation because traffic anomalies in roundabout scenes are often subtle and complex, making it difficult even for strong commercial MLLMs to capture all details reliably. These human-labelled annotations are then used as prior context for the following stages. In \textit{Stage 2}, we use GPT-5 \cite{openai_gpt5_2025_08_07} to enrich the annotations and generate structured labels for multiple aspects of the event, including: (1) environment information; (2) anomaly object grounding; (3) event information; and (4) a refined event summary. In \textit{Stage 3}, we further manually inspect and correct inaccurate machine-generated annotations to ensure label quality. In \textit{Stage 4}, we again use GPT-5 to refine and enrich the final event summary, using the verified decomposed annotations as grounding information so that the summary consistently covers environment, object grounding, event description, and event analysis. Following this pipeline, we finally construct 2,064 QA pairs.
\subsection{Dataset statistics}
\begin{figure}[t]
  \centering
  \includegraphics[width=\textwidth]{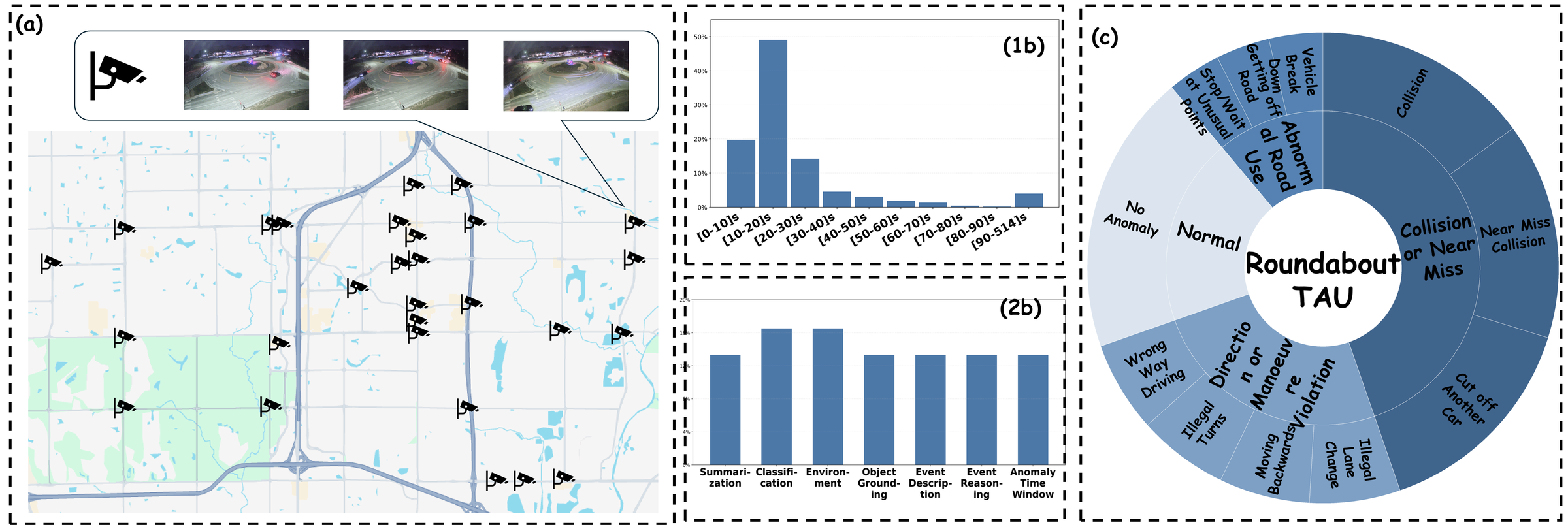}
  \caption{\textbf{Roundabout-TAU dataset statistics.} (a) Overview of Roundabout-TAU, including the map of 28 camera sites and one example video. (1b) Distribution of video lengths in Roundabout-TAU. (2b) Proportion of different QA categories in Roundabout-TAU. (c) Distribution of anomaly classes.}
  \label{fig:dataset}
\end{figure}
Roundabout-TAU is a real-world traffic scene dataset designed for traffic anomaly understanding, as illustrated in \cref{fig:dataset}. It contains 342 videos collected from 28 different sites, covering diverse camera viewpoints and real-world traffic conditions, as shown in \cref{fig:dataset}(a). In total, the dataset includes around 150 minutes of video, with clips of varying lengths, as shown in \cref{fig:dataset}(1b). To support anomaly classification, we group the videos into four anomaly categories representing different types of traffic events, as shown in \cref{fig:dataset}(c). To support deeper traffic anomaly understanding, we further annotate the videos with multiple QA categories that cover different aspects of the scene and event, as shown in \cref{fig:dataset}(2b).

\section{Method: TAU-R1}
In this section, we propose TAU-R1, a novel two-layer hierarchical framework for traffic anomaly understanding. TAU-R1 is designed to support efficient inference on real-world edge-deployed systems. To further improve the model's reasoning ability for traffic anomaly understanding, we introduce a two-stage training strategy: (1) decomposed-QA enhanced supervised fine-tuning (SFT) and (2) TAU-GRPO, a task-driven reinforcement learning based post-training method for traffic anomaly understanding.
\subsection{TAU-R1 Framework}
\begin{figure}[t]
  \centering
  \includegraphics[width=\textwidth]{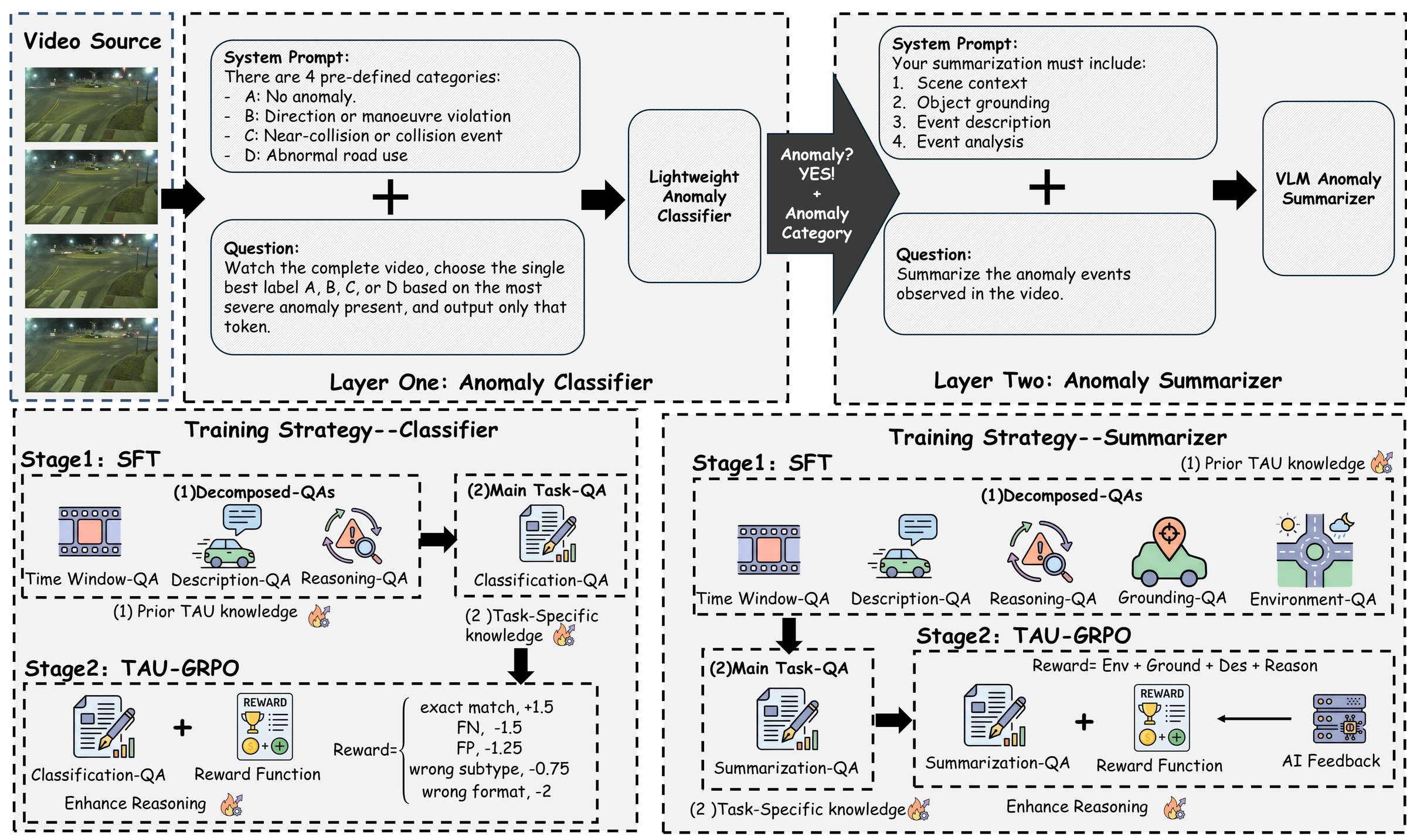}
    \caption{TAU-R1 framework and training pipeline. \textbf{Top:} the two-layer framework, where a lightweight classifier filters the video stream and a larger reasoner summarises anomalous clips. \textbf{Bottom:} the two-stage training strategy for both modules, including decomposed-QA enhanced SFT followed by TAU-GRPO post-training.}
  \label{fig:tau-r1}
\end{figure}
TAU-R1 is a two-layer hierarchical framework for traffic anomaly understanding, consisting of (1) an anomaly classifier and (2) an anomaly summariser. The overall structure is shown in \cref{fig:tau-r1}. We design this framework for real-world deployment on resource-constrained edge devices, where continuous traffic monitoring requires efficient and stable inference. Under such constraints, only lightweight VLMs, typically smaller than 8B parameters \cite{guan2025roadscenevqa}, are practical candidates for real-time deployment.

Based on this observation, we assign different subtasks to models with different capabilities. In the first layer, a lightweight model performs anomaly classification, which requires less reasoning and supports fast inference. This layer provides an initial judgment by classifying the input video into one of four predefined categories. If the event is classified as anomalous, the raw video together with the predicted anomaly class is passed to the second layer. In the second layer, a larger model performs anomaly summarisation, leveraging stronger reasoning ability to generate a detailed understanding of the event.
\subsection{Decomposed-QA Enhanced SFT}
When humans perceive a traffic anomaly, they typically make a direct judgement as the anomaly occurs, or slightly before it occurs, rather than reasoning sequentially. We argue that this is because humans are very familiar with traffic scenes and have a good prior understanding of the traffic environment, traffic rules, typical vehicle behaviour, and the interactions among road users.

\begin{figure}[t]
  \centering
  \includegraphics[width=\textwidth]{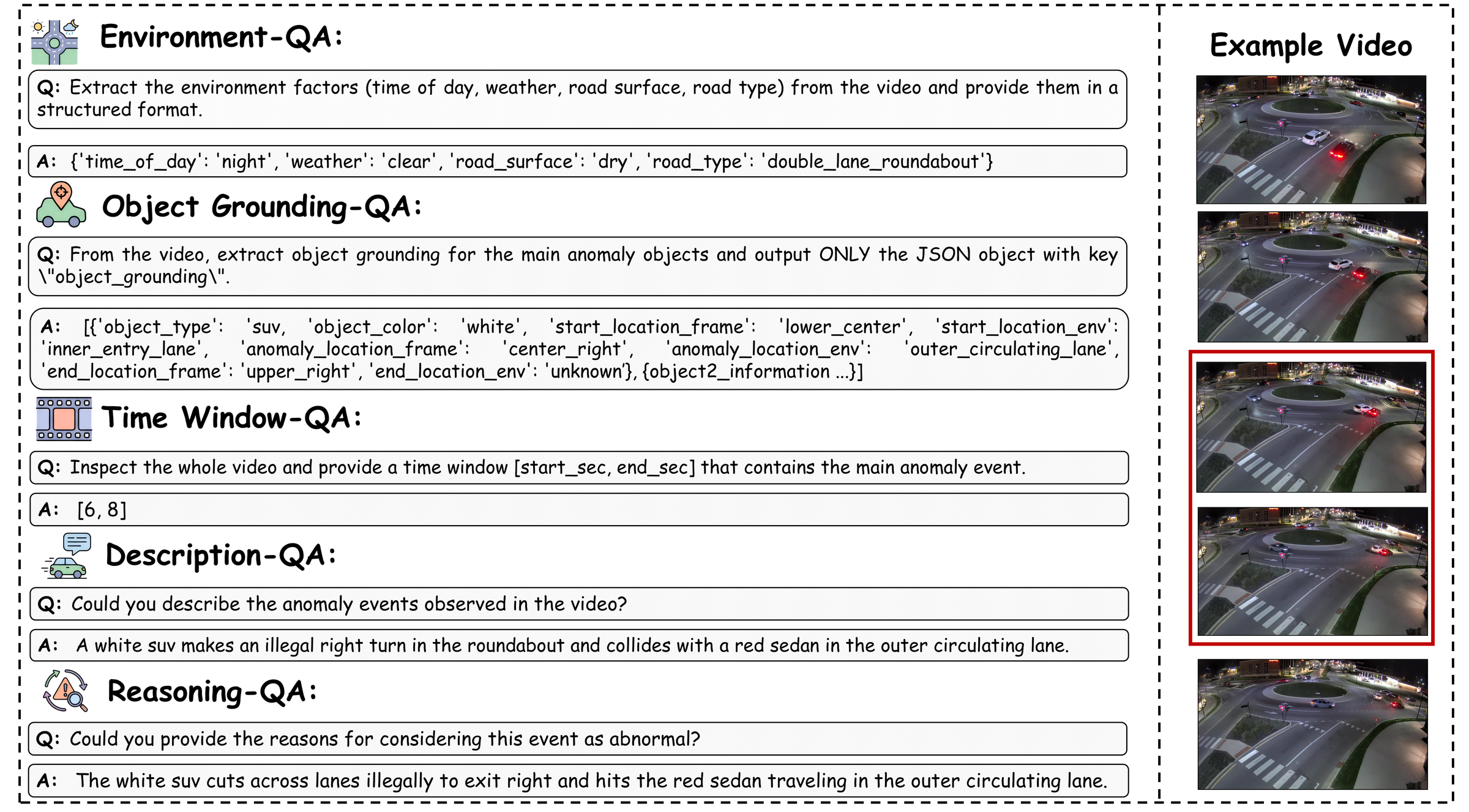}
  \caption{\textbf{Decomposed-QA enhanced SFT.} To help the model learn the intermediate scene knowledge required for traffic anomaly understanding, we decompose the task into five question-answer types: environment QA, object grounding QA, anomaly time window QA, anomaly reasoning QA, and anomaly description QA. The figure shows one example for each question type.}
  \label{fig:decompose-qa}
\end{figure}

Motivated by this observation, we argue that directly asking a VLM to generate anomaly understanding outputs without explicitly learning these priors is difficult. Therefore, instead of training the model only on the final tasks, we decompose traffic anomaly understanding into five question-answer tasks: environment QA, object grounding QA, anomaly time window QA, anomaly reasoning QA, and anomaly description QA. We further illustrate these five question types with one example each in \cref{fig:decompose-qa}. Overall, this design encourages the model to learn the intermediate scene knowledge required for anomaly reasoning, thereby improving its final anomaly understanding performance.

\subsection{TAU-GRPO}
To further improve reasoning for traffic anomaly understanding, we apply reinforcement learning based post-training on top of SFT. We adopt Group Relative Policy Optimisation (GRPO) \cite{shao2024deepseekmath} for its strong reasoning gains and efficient group-based optimisation. However, directly applying vanilla GRPO to TAU is suboptimal, because TAU includes both structured classification and open-ended summarisation that cannot be evaluated by simple binary correctness. Therefore, we propose \textbf{TAU-GRPO}, a TAU-oriented extension of GRPO with customised reward functions for anomaly classification and summarisation.
\subsubsection{TAU-GRPO Reward Function}
For TAU, we design separate reward functions for anomaly classification and anomaly summarisation, since the two tasks require different forms of supervision.

For anomaly classification, we design the reward according to four objectives: (1) improving four-class classification accuracy, (2) improving binary classification (normal/abnormal) accuracy, (3) penalising false negatives more heavily than false positives because missing an abnormal event is typically more costly than incorrectly flagging a normal event as abnormal in real-world deployment, and (4) enforcing valid output formatting. Let \(y\) and \(\hat{y}\) denote the ground-truth and predicted labels, respectively, and let \(A\) denote the normal class while \(B\), \(C\), and \(D\) denote different abnormal classes. The classification reward is defined as
\begin{equation}
R_{\mathrm{cls}}(\hat{y},y)=
\begin{cases}
-2.0, & \hat{y}\notin\{A,B,C,D\} \ \text{invalid format}, \\
-1.50, & y\in\{B,C,D\},\ \hat{y}=A \ \text{false negative}, \\
-1.25, & y=A,\ \hat{y}\in\{B,C,D\} \ \text{false positive}, \\
-0.75, & y,\hat{y}\in\{B,C,D\},\ \hat{y}\neq y \ \text{wrong abnormal subtype}, \\
1.5, & \hat{y}=y \ \text{exact class match}.
\end{cases}
\end{equation}

For anomaly summarisation, traditional natural language processing metrics such as BLEU \cite{papineni2002bleu} and METEOR \cite{banerjee2005meteor} are insufficient, since they mainly reflect surface text-level similarity rather than the actual quality of anomaly understanding. To provide a reward that better matches the TAU task, we follow the Reinforcement Learning from AI Feedback (RLAIF) paradigm \cite{bai2022constitutional}, where an LLM judge evaluates the generated summary against the ground-truth summary from multiple aspects:
\begin{equation}
R_{\mathrm{sum}}(\hat{y},y)
=
R_{\mathrm{env}}
+
R_{\mathrm{ground}}
+
R_{\mathrm{desc}}
+
R_{\mathrm{reason}}
-
P_{\mathrm{hall}}
-
P_{\mathrm{verb}}.
\end{equation}
Specifically, \(R_{\mathrm{env}} \in [0,1]\) evaluates environment correctness, \(R_{\mathrm{ground}} \in [0,2]\) evaluates object grounding quality, \(R_{\mathrm{desc}} \in [0,5]\) evaluates event description quality, and \(R_{\mathrm{reason}} \in [0,2]\) evaluates anomaly reasoning quality. The penalty terms \(P_{\mathrm{hall}} \in [0,3]\) and \(P_{\mathrm{verb}} \in [0,1]\) discourage hallucinated content and unnecessary verbosity, respectively. We assign the largest weight to \(R_{\mathrm{desc}}\) because event description captures the core anomaly content in the summary. Overall, this design aligns the reward more closely with the actual objectives of TAU summarisation and improves the reasoning capability of our model.
\subsection{Training Pipeline}
For TAU-R1, we adopt a two-stage training pipeline for both the anomaly classifier and summariser, as illustrated in \cref{fig:tau-r1}. The design follows the intuition introduced in the decomposed-QA section: traffic anomaly understanding should first acquire prior context, rather than learning only from the final task output.

\subsubsection{Stage 1: Supervised Fine-Tuning}
In Stage 1, we perform full-parameter supervised fine-tuning on decomposed QA tasks to help the model learn prior knowledge for traffic anomaly understanding. For the anomaly classifier, we train on anomaly time-window QA, anomaly description QA, and anomaly reasoning QA for three epochs. These tasks provide the most relevant priors for classification while avoiding less essential supervision for a lightweight model. We then fine-tune the classifier on the anomaly classification QA set for six epochs to adapt it to the target labels and output format.

For the anomaly summariser, we follow a similar strategy but use all decomposed QA tasks, including environment QA, object grounding QA, anomaly time-window QA, anomaly description QA, and anomaly reasoning QA, for three epochs. Since summarisation requires richer scene understanding and stronger reasoning, we then fine-tune the model on the summarisation QA set for four epochs to adapt it to the final generation task and response style.

\subsubsection{Stage 2: GRPO-based Post-Training} In Stage 2, we further conduct full-parameter TAU-GRPO post-training on both models on their respective final-task datasets for one epoch, in order to enhance task-specific reasoning and improve alignment with the target objectives.
\section{Experiment}
\subsection{Experimental Settings}
\subsubsection{Datasets}
The Roundabout-TAU dataset contains 342 video clips in total, with anomaly class distributions shown in \cref{fig:dataset}(c). For evaluation, we construct a test set of 42 videos by random sampling while maintaining the overall class distribution of the full dataset. The remaining 300 videos are used for training. We report anomaly classification performance on all 42 test videos. For anomaly summarisation, we evaluate only the anomalous samples in the test set, since normal videos do not require anomaly summaries. This results in 34 videos used for summarisation evaluation.
\subsubsection{Evaluation Metrics}
We evaluate TAU-R1 on the two tasks defined in Roundabout-TAU: anomaly classification and anomaly summarisation. For classification, we report both four-class and binary classification performance using Average Precision (AP) and F1 score. For summarisation, following prior video anomaly understanding work \cite{zhang2025holmes,xing2025echotraffic}, we use standard text-generation metrics including BLEU \cite{papineni2002bleu}, METEOR \cite{banerjee2005meteor} and ROUGE-L \cite{lin2004rouge}. In addition, we introduce \textbf{LLM-Score} to evaluate semantic accuracy and reasoning quality, following recent LLM-based evaluation practices \cite{liu2023visual,tang2024hawk}. Specifically, we use DeepSeek-V4-Pro, Claude Opus 4.6, and Gemini 3 Flash as judges. Each model independently scores a prediction from 0 to 10 based on four aspects: environment correctness (0--1), object grounding (0--2), description quality (0--5), and reasoning quality (0--2). The final LLM-Score is the average score across the three models.
\subsubsection{Implementation Detail}
We build TAU-R1 based on Qwen3-VL~\cite{Qwen3-VL}. For the lightweight first layer, we use Qwen3-VL-2B, while for the larger second layer, we use Qwen3-VL-8B. Model training and the main task performance evaluation are conducted on two NVIDIA A40 GPUs. The inference efficiency and deployment experiments are conducted on an NVIDIA Jetson AGX Orin (64\,GB). 
\subsection{Main Task Performance}
We compare classification and summarisation results against multiple state-of-the-art baselines: open-source MLLMs Qwen3-VL \cite{Qwen3-VL} and Intern-VL3 \cite{zhu2025internvl3}; open-source video reasoning MLLMs Video-LLaVA \cite{lin2024video} and Video-R1 \cite{feng2025video}; video anomaly specific MLLMs Hawk \cite{tang2024hawk}, VAD-R1 \cite{huang2025vad} and EchoTraffic \cite{xing2025echotraffic}; and commercial MLLM APIs GPT-5 \cite{openai_gpt5_2025_08_07}, Qwen3.5-Plus \cite{qwen3.5} and Gemini-3.1-pro \cite{google_gemini_2026}. The results are shown in \cref{tab:main_results}.

\begin{table*}[t]
\centering
\captionsetup[table]{skip=5pt}
\vspace{-8pt}
\caption{Comparison of different approaches on the task of traffic anomaly understanding over Roundabout-TAU dataset. \textbf{Bold} values indicate the best performance, while \underline{underlined values} indicate the second-best performance. * denotes fine-tuning on the Roundabout-TAU training set.}
\label{tab:main_results}
\resizebox{\textwidth}{!}{%
\begin{tabular}{lccccccccc}
\toprule
\multirow{2}{*}{\textbf{Model}} & \multirow{2}{*}{\textbf{Params.}} & \multicolumn{4}{c}{\textbf{Classification}} & \multicolumn{4}{c}{\textbf{Summarisation}} \\
\cmidrule(lr){3-6} \cmidrule(lr){7-10}
& & 4-cls AP & 4-cls F1 & 2-cls AP & 2-cls F1 & BLEU & ROUGE-L & METEOR & LLM-Score \\
\midrule
\rowcolor{gray!20}
\multicolumn{10}{c}{\textit{Commercial MLLMs}} \\
\midrule
GPT-5 \cite{openai_gpt5_2025_08_07}          & N/A   & 0.2143 & 0.1115 & 0.2857 & 0.2367 & 0.0353 & \underline{0.2539} & \underline{0.2842} & \underline{3.2517} \\
Qwen3.5-Plus \cite{qwen3.5}   & N/A   & 0.2143 & 0.1074 & 0.2143 & 0.1085 & 0.0313 & 0.2295 & 0.2801 & 2.4813 \\
Gemini-3.1-pro \cite{google_gemini_2026} & N/A   & \underline{0.4048} & \underline{0.4086} & 0.5952 & 0.6372 & 0.0417 & 0.2423 & 0.2660 & 2.9392 \\
\midrule
\rowcolor{gray!20}
\multicolumn{10}{c}{\textit{Open-Source MLLMs}} \\
\midrule
Qwen3-VL \cite{Qwen3-VL}       & 8B    & 0.2381 & 0.1967 & 0.6190 & 0.6452 & 0.0325 & 0.2534 & 0.2756 & 3.0230 \\
Intern-VL3 \cite{zhu2025internvl3}     & 8B    & 0.1905 & 0.1008 & 0.1905 & 0.1005 & 0.0349 & 0.2367 & 0.2686 & 2.4428 \\
\midrule
\rowcolor{gray!20}
\multicolumn{10}{c}{\textit{Open-Source Video Reasoning MLLMs}} \\
\midrule
Video-LLaVA \cite{lin2024video}    & 7B    & 0.1667 & 0.0606 & 0.1905 & 0.0610 & 0.0192 & 0.1825 & 0.1858 & 1.2185 \\
Video-R1 \cite{feng2025video}       & 7B    & 0.2143 & 0.1087 & 0.2381 & 0.1534 & 0.0198 & 0.2108 & 0.2333 & 1.3382 \\
\midrule
\rowcolor{gray!20}
\multicolumn{10}{c}{\textit{Video Anomaly Specific MLLMs}} \\
\midrule
Hawk \cite{tang2024hawk}           & 7B    & 0.0952 & 0.0938 & 0.5952 & 0.6274 & 0.0118 & 0.1596 & 0.2154 & 0.4967 \\
VAD-R1 \cite{huang2025vad}         & 7B    & 0.2381 & 0.1345 & \underline{0.7619} & \underline{0.7001} & \underline{0.1331} & 0.2463 & 0.1788 & 2.0357 \\
EchoTraffic \cite{xing2025echotraffic}    & 7B    & N/A    & N/A    & N/A    & N/A    & 0.0000 & 0.1877 & 0.1958 & 0.8561 \\
\midrule
\rowcolor{gray!20}
\multicolumn{10}{c}{\textit{Roundabout-TAU Fine-Tuned MLLMs}} \\
\midrule
Qwen3-VL* \cite{Qwen3-VL}       & 8B  &  0.5476 & 0.5121 & 0.8810 & 0.8580 & 0.1749 & 0.4050 & 0.4104 & 3.9985 \\
VAD-R1* \cite{huang2025vad}  & 7B & 0.5238 & 0.4352 & 0.8333 & 0.7763 & 0.1489 & 0.3678 & 0.3592 & 2.9709 \\
\midrule
\textbf{TAU-R1 (Ours)}   & 2B/8B & \textbf{0.7381} & \textbf{0.7310} & \textbf{0.9286} & \textbf{0.9218} & \textbf{0.2091} & \textbf{0.4234} & \textbf{0.4252} & \textbf{4.6588} \\
\bottomrule
\end{tabular}%
}
\end{table*}

As shown in \cref{tab:main_results}, TAU-R1 achieves the strongest overall performance on both the classification and the summarisation task, outperforming all listed state-of-the-art methods as well as the fine-tuned open-source baselines. This demonstrates the effectiveness of our two-layer framework and task-specific training strategy for traffic anomaly understanding. We also observe that most existing methods perform poorly on Roundabout-TAU, indicating that real-world roadside roundabout scenes remain challenging for current models. Overall, these results validate our proposed methodology while also highlighting the need for benchmarks and training strategies specifically designed for roadside traffic anomaly understanding.

\subsection{Inference Efficiency and Deployment Performance}
We evaluate inference efficiency on a resource-constrained edge device using an NVIDIA Jetson AGX Orin (64\,GB). We deploy the two-layer framework and measure inference time on Roundabout-TAU. Specifically, the classifier processes all 342 videos (approximately 150 minutes), and the summariser is applied to the 276 anomalous videos (approximately 130 minutes). The results are reported in \cref{tab:jetson_efficiency}.

\begin{table}[t]
\centering
\captionsetup[table]{skip=5pt}
\vspace{-8pt}
\caption{Jetson AGX Orin deployment efficiency on Roundabout-TAU.}
\label{tab:jetson_efficiency}
\resizebox{\textwidth}{!}{
\begin{tabular}{lccc}
\toprule
\textbf{Metric} & \textbf{Classifier} & \textbf{Summariser} &\textbf{End-to-End} \\
\midrule
Total runtime (sec) & 809.13 & 3441.92 & 4251.05 \\
Avg latency per clip (sec/clip) & 2.37 & 12.47 & 12.43 \\
Real-Time Factor (runtime/video\_time) & 0.09 & 0.44 & 0.47 \\
\bottomrule
\end{tabular}}
\end{table}

Overall, the lightweight classifier requires only 2.37 seconds per clip, making it suitable as a first-stage filter. The end-to-end pipeline achieves a real-time factor (RTF) of 0.47, i.e.\ 2.1$\times$ faster than real time. These results validate the advantage of our hierarchical design: the majority of videos can be handled by the low-cost classifier, while the more expensive summariser is invoked only when an anomaly is detected. This makes our framework promising for real-world TAU deployment, and also suggests feasibility on even more cost-efficient devices (e.g.\ Orin Nano) with appropriate model and precision settings.

\subsection{Ablation Study}
We ablate the proposed training strategy by progressively adding its components (\cref{tab:ablation}). Overall, both anomaly classification and anomaly summarisation improve consistently. For classification, all major metrics increase step by step, and TAU-GRPO further reduces binary false negatives from 2 to 0, consistent with our reward design that encourages classifying uncertain cases as abnormal rather than missing potentially anomalous events as the latter are more costly in real-world deployment. For summarisation, the SFT stages also improve performance consistently, while TAU-GRPO mainly improves LLM-Score rather than BLEU/METEOR, which is expected because TAU-GRPO targets reasoning quality instead of text-based overlap.

\begin{table}[t]
\centering
\captionsetup[table]{skip=5pt}
\vspace{-8pt}
\caption{Ablation study of the proposed training pipeline on the Roundabout-TAU dataset. The classification ablation is performed on the 2B model, whereas the summarisation ablation is performed on the 8B model. FP and FN denote the false positives and false negatives of binary classification, respectively.}
\label{tab:ablation}
\resizebox{\linewidth}{!}{%
\begin{tabular}{lcccccccccc}
\toprule
\multirow{2}{*}{Method} & \multicolumn{6}{c}{Classification} & \multicolumn{4}{c}{Summarisation} \\
\cmidrule(lr){2-7} \cmidrule(lr){8-11}
& 4-cls AP & 4-cls F1 & 2-cls AP & 2-cls F1 & FP & FN & BLEU & ROUGE-L & METEOR & LLM-Score \\
\midrule
Qwen3-VL            & 0.2381 & 0.1967 & 0.6190 & 0.6452 & 6 & 10 & 0.0325 & 0.2534 & 0.2756 & 3.0230 \\
+SFT                 & 0.5476 & 0.5121 & 0.8810 & 0.8580 & 5 & 0 & 0.1749 & 0.4050 & 0.4104 & 3.9985 \\
+Decomposed-QA SFT  & 0.6429 & 0.6400 & 0.9048 & 0.9048 & \textbf{2} & 2 & \textbf{0.2110} & 0.4171 & \textbf{0.4298} & 4.2220 \\
+TAU-GRPO           & \textbf{0.7381} & \textbf{0.7310} & \textbf{0.9286} & \textbf{0.9218} & 3 & \textbf{0} & 0.2091 & \textbf{0.4234} & 0.4252 & \textbf{4.6588} \\
\bottomrule
\end{tabular}%
}
\end{table}
\subsection{Split Robustness Analysis}
To assess the robustness of our results and mitigate concerns regarding the relatively small test set, we re-construct an additional train-test split that is disjoint from the original evaluation set. Specifically, the 42 videos used in the original test set are excluded from the new test split, allowing evaluation on a different set of previously unseen videos. The new test set contains 42 videos, matching the size of the original test set, and the whole framework was retrained from scratch for this experiment. As shown in \cref{tab:split_robustness}, the overall performance trend remains consistent with the original split: each stage of the training pipeline yields incremental improvements, while TAU-GRPO achieves the best performance in both four-class anomaly classification and LLM-Score. These results indicate that the effectiveness of the proposed approach is not tied to a particular dataset split.

\begin{table}[t]
\centering
\caption{Split robustness analysis on a disjoint train--test split.}
\captionsetup[table]{skip=5pt}
\vspace{-8pt}
\label{tab:split_robustness}
\resizebox{\linewidth}{!}{%
\begin{tabular}{lcccccccccc}
\toprule
\multirow{2}{*}{Method} & \multicolumn{6}{c}{Classification} & \multicolumn{4}{c}{Summarisation} \\
\cmidrule(lr){2-7} \cmidrule(lr){8-11}
& 4-cls AP & 4-cls F1 & 2-cls AP & 2-cls F1 & FP & FN & BLEU & ROUGE-L & METEOR & LLM-Score \\
\midrule
Qwen3-VL  & 0.2142 & 0.1601 & 0.5476 & 0.5945& 5 & 14 & 0.0405 & 0.2378 & 0.2605 & 2.5552 \\
+SFT      & 0.6429 & 0.5715 & \textbf{0.9762} & \textbf{0.9756} & \textbf{1} & \textbf{0} & 0.1542 & 0.3840 & 0.3585 & 3.6858 \\
+Decomposed-QA-SFT  & 0.7142 & 0.7113 & 0.9524 & 0.9524 & 1 & 1 & \textbf{0.1751} & \textbf{0.4164} & \textbf{0.3886} & 4.6442 \\
+TAU-GRPO & \textbf{0.8095} & \textbf{0.8014} & 0.9524 &0.9524 & 1 & 1 & 0.1593 & 0.4000 & 0.3646 & \textbf{4.7963} \\
\bottomrule
\end{tabular}%
}
\end{table}

\subsection{Out-of-domain Performance}
To examine the out-of-domain performance of TAU-R1, we selected 15 non-roundabout traffic anomaly videos from the City of Carmel dataset, annotated using the same labelling protocol as our in-domain data. The quantitative results are presented in \cref{tab:ood-evaluation}. On this small 15-video out-of-domain evaluation, TAU-R1 achieves the best result across the reported metrics. These preliminary results suggest knowledge is transferred beyond roundabout scenes, although a larger and more diverse evaluation is required to establish robust domain generalisation.

\begin{table}[t]
\centering
\caption{Generalisation evaluation on out-of-domain data.}
\captionsetup[table]{skip=5pt}
\vspace{-8pt}
\label{tab:ood-evaluation}
\footnotesize
\begin{tabular}{lcccc}
\toprule
\textbf{Model} & \textbf{BLEU} & \textbf{ROUGE-L} & \textbf{METEOR} & \textbf{LLM-Score} \\
\midrule
Qwen3-VL  & 0.0206 & 0.2027 & 0.2543 & 2.9277 \\
EchoTraffic & 0.0000 & 0.1780 & 0.2081 & 1.4627 \\
VAD-R1      & 0.0082 & 0.1482 & 0.2172 & 1.9998 \\
\textbf{TAU-R1 (Ours)} & \textbf{0.0226} & \textbf{0.2421} & \textbf{0.2594} & \textbf{2.9538} \\
\bottomrule
\end{tabular}
\end{table}

\subsection{Qualitative Comparison}
\begin{figure}[t]
  \centering
  \includegraphics[width=\textwidth]{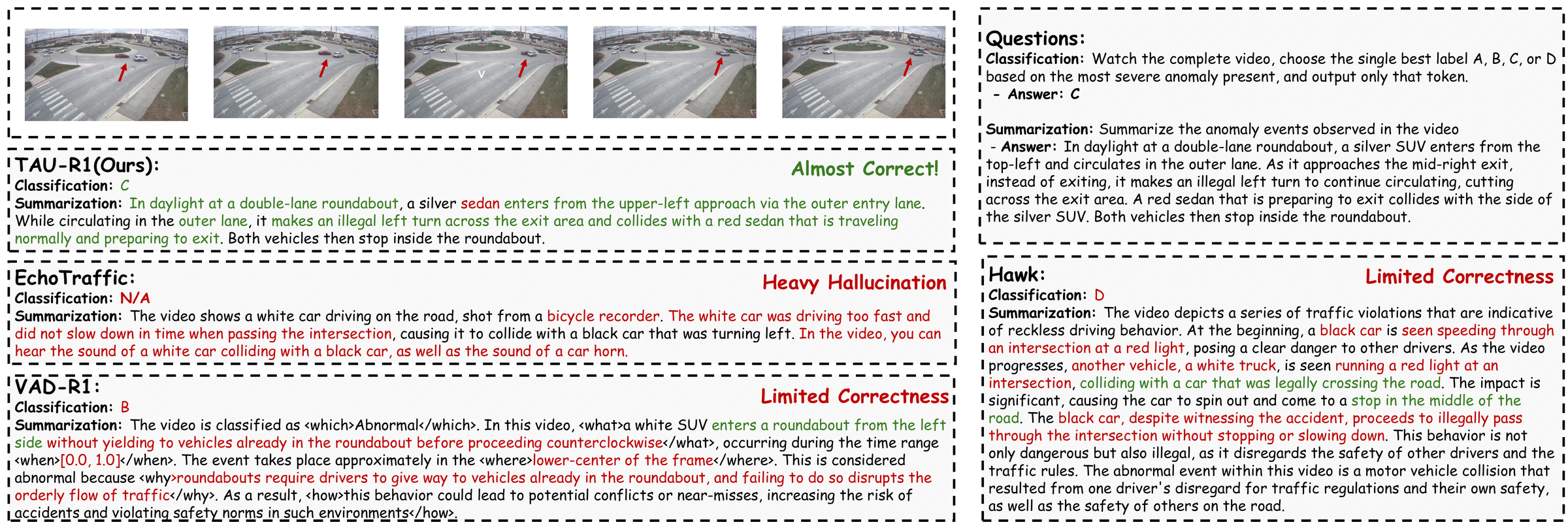}
  \caption{Qualitative comparison on the Roundabout-TAU dataset.}
  \label{fig:qualititive_analysis}
\end{figure}

We qualitatively compare TAU-R1 with other video-anomaly-specific MLLMs on Roundabout-TAU in \cref{fig:qualititive_analysis}. Overall, TAU-R1 gives more accurate outputs for both classification and summarisation. Compared to the baselines, TAU-R1 can capture and understand the key anomaly event with very limited hallucination, while other models more or less hallucinate details and sometimes fail to catch the key anomaly event.
\section{Conclusion}
In conclusion, we introduce Roundabout-TAU, a real-world roadside benchmark for traffic anomaly understanding built from roundabout surveillance videos, with 342 clips and 2,000+ multi-aspect QA pairs enabling fine-grained evaluation beyond coarse anomaly labels. We further propose TAU-R1, a deployment-oriented two-layer VLM framework trained via decomposed-QA enhanced SFT and TAU-GRPO post-training, which achieves the best overall performance on both anomaly classification and anomaly summarisation among diverse baselines, while remaining practical for edge deployment on Jetson AGX Orin. We hope Roundabout-TAU and TAU-R1 will facilitate future research on reliable and interpretable VLM-based traffic safety systems for real-world deployment.

\section*{Acknowledgements}
Yuqiang Lin and Sam Lockyer are supported by a scholarship from the EPSRC Centre for Doctoral Training in Advanced Automotive Propulsion Systems (AAPS) under project EP/S023364/1.

% ---- Bibliography ----
\bibliographystyle{splncs04}
\bibliography{main}
\FloatBarrier
\newpage
\appendix
\section{Overview of the Appendix}
This appendix provides supplementary material that could not be included in the main paper due to space limitations. First, we provide additional details about the proposed Roundabout-TAU dataset, including a comparison with existing video anomaly detection and video anomaly understanding datasets, privacy considerations for the video sources, and the detailed prompt templates used to construct the QA pairs. Second, we provide additional experimental details to improve reproducibility, including training and inference settings and the full prompt templates for LLM-based evaluation and the TAU-GRPO summarization reward function. Finally, we present additional qualitative results on both in-domain and out-of-domain traffic scenes, followed by a discussion of limitations and future research directions.
\section{Roundabout-TAU Dataset}
\subsection{Dataset Comparison}
In this section, we compare Roundabout-TAU with existing open-source VAD and VAU datasets collected from real-world videos, as shown in \cref{tab:dataset_comparison}. Since many existing VAU datasets are mainly based on web-mined videos, we focus here on datasets collected from real-world settings to better highlight the data collection contribution of Roundabout-TAU.

Roundabout-TAU contains 342 video clips, including 276 anomalous clips, collected from 28 fixed roadside cameras, with a total duration of over 2.5 hours. As shown in \cref{tab:dataset_comparison}, Roundabout-TAU contains the largest number of anomalous clips among the open-source datasets listed here. It also covers a relatively diverse set of sites, second only to the MSAD dataset~\cite{MSAD}. In addition, the dataset maintains a practical total video duration for traffic anomaly analysis. More importantly, unlike these prior datasets, Roundabout-TAU additionally provides text-based annotations, making it suitable not only for anomaly detection but also for video-language-model training and traffic anomaly understanding tasks.

Overall, Roundabout-TAU helps fill an important gap in traffic anomaly research, particularly the lack of real-world roadside traffic videos captured by fixed cameras. This dataset is the result of a dedicated two-year effort to collect and organize real-world traffic anomaly videos.
\begin{table}[h]
\centering
\caption{Comparison of video anomaly detection/understanding datasets using videos collected from real-world settings.}
\label{tab:dataset_comparison}
\resizebox{\textwidth}{!}{
\begin{tabular}{lcccccc}
\toprule
\textbf{Dataset} & \textbf{Sites} & \textbf{Scenes} & \textbf{Anom. Clips} & \textbf{Total Clips} & \textbf{Hours} & \textbf{Text Ann.} \\
\midrule
ShanghaiTech (CVPR'18)~\cite{shanghaitech}   & 13  & Campus   & 130     & 437     & 3.7h     & $\times$ \\
UCSD-Ped (TPAMI'13)~\cite{ucsd-ped}       & 2   & Campus   & 48      & 98      & 0.5h     & $\times$ \\
CUHK Avenue (ICCV'13)~\cite{cuhk}    & 1   & Campus   & 21      & 37      & 0.3h    & $\times$ \\
MSAD (NeurIPS'24)~\cite{MSAD}           & 500 & Multiple & 240     & 720     & \textbf{4.1h}     & $\times$ \\
\midrule
\textbf{Roundabout-TAU(Ours)} & 28  & Traffic  & \textbf{276}     & 342     & 2.5h    & \checkmark \\
\bottomrule
\end{tabular}
}
\end{table}

\subsection{Data Privacy}
Privacy protection was considered throughout the construction of Roundabout-TAU. The videos were collected from fixed roadside cameras with relatively long viewing distances and side-angled perspectives, which naturally limit the visibility of personally identifiable information. In particular, human faces and vehicle number plates are generally unrecognizable in the recorded footage. To further minimize privacy risks, we manually reviewed the videos and confirmed that the released data contained no clearly recognizable faces or readable number plates. Therefore, the dataset preserves the traffic scene information required for anomaly understanding while avoiding the exposure of directly identifiable personal details.

\subsection{Detailed Prompts for Dataset Construction}
\begin{figure}[t!]
  \centering
  \includegraphics[width=0.95\textwidth]{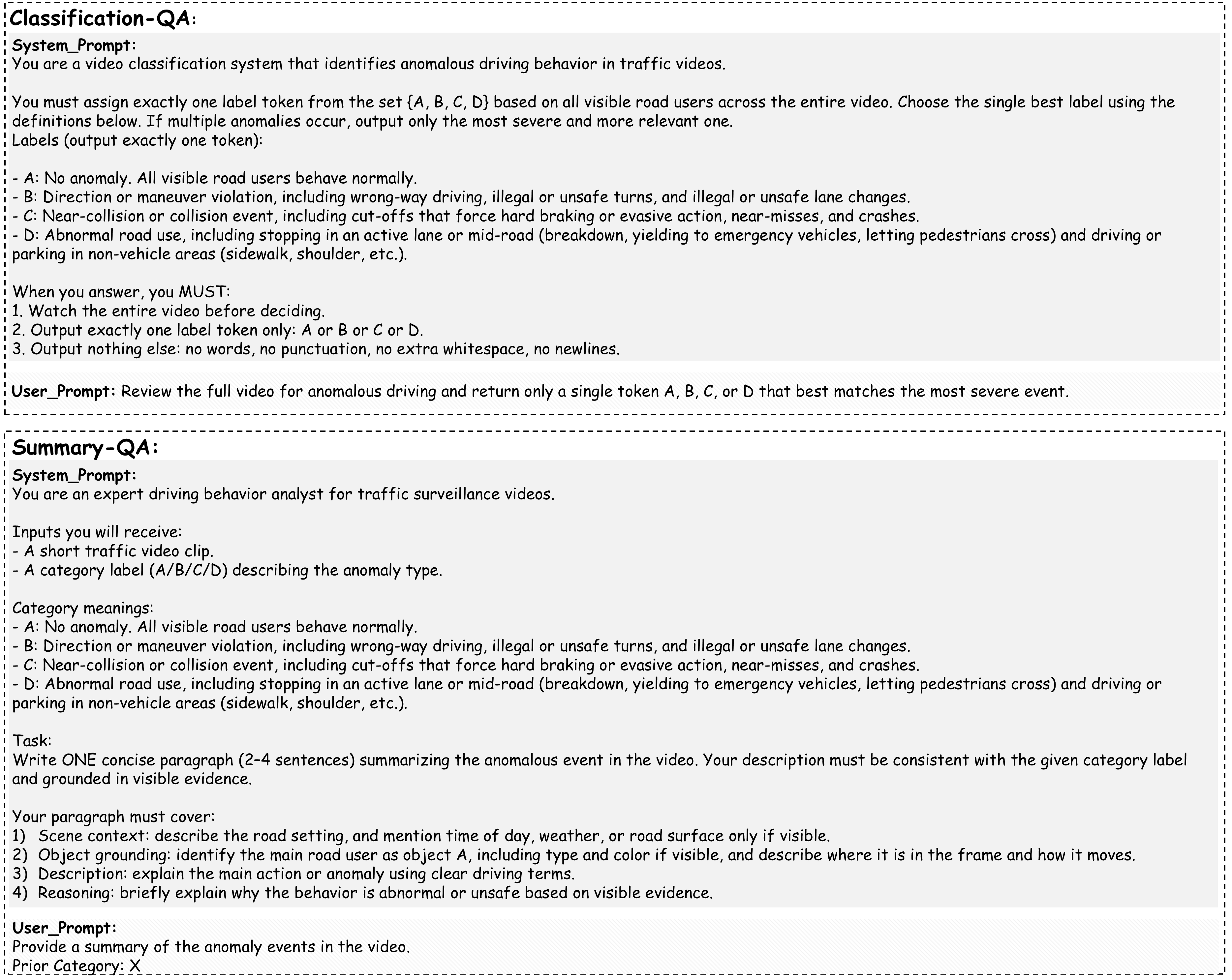}
  \caption{Prompt templates for the classification and summarization tasks.}
  \label{fig:appendix_main_tasks}
\end{figure}

\begin{figure}[t!]
  \centering
  \includegraphics[width=0.95\textwidth]{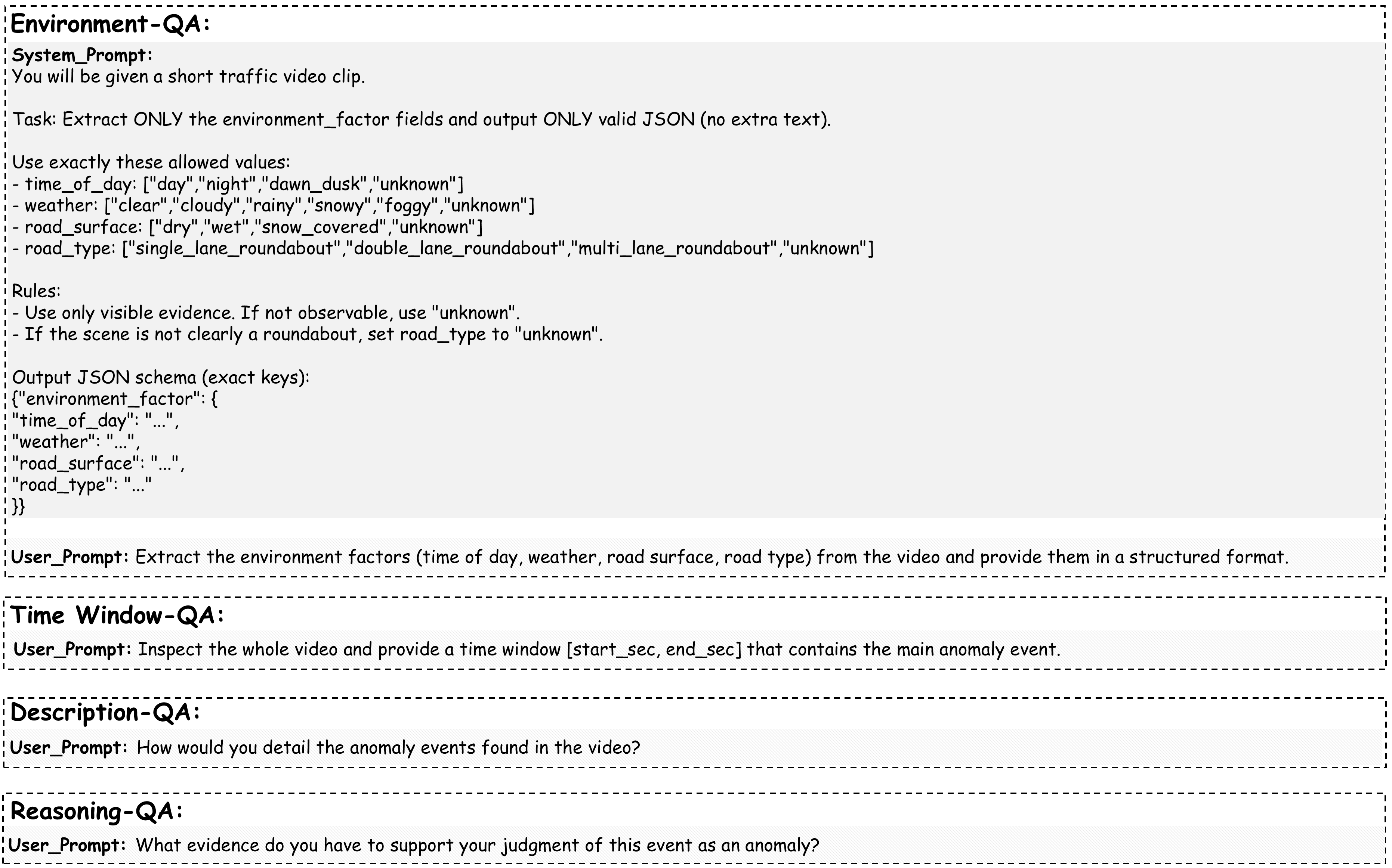}
  \caption{Prompt templates for the other decomposed-QA tasks.}
  \label{fig:appendix_other_decompose_qa}
\end{figure}

In this section, we present the full prompt templates used to construct and organize the QA pairs in Roundabout-TAU, with the aim of improving the reproducibility of our dataset construction pipeline. Specifically, we include the prompt templates for the two main tasks, anomaly classification and anomaly summarization, in \cref{fig:appendix_main_tasks}, as well as the complete prompt templates for all decomposed QA tasks in \cref{fig:appendix_grounding_qa,fig:appendix_other_decompose_qa}.

\begin{figure}[t!]
  \centering
  \includegraphics[width=0.95\textwidth]{fig/grounding-qa.png}
  \caption{Prompt template for the object grounding task.}
  \label{fig:appendix_grounding_qa}
\end{figure}
% ---------------------------
\section{More Experimental Details}
\subsection{Training and Inference Details}
All experiments were conducted on two NVIDIA A40 GPUs. During the supervised fine-tuning (SFT) stage, including the decomposed-QA training phase, the learning rate was set to \(5\times10^{-6}\) for the anomaly classifier and \(3\times10^{-6}\) for the anomaly summarizer. During the second-stage TAU-GRPO training, both models used a learning rate of \(1\times10^{-6}\). Across all training stages, the per-device batch size was set to 1, with gradient accumulation over 4 steps. The complete training pipeline for the two models takes approximately 27.5 hours.

To improve memory efficiency, we adopted DeepSpeed ZeRO-3 with CPU offloading. For video input processing, we used a frame sampling rate of 2 FPS, with 32--256 visual tokens per frame, a maximum of \(1.28\times10^4\) visual tokens per video, and a maximum of 180 input frames. These constraints were applied jointly, with the video inputs dynamically adjusted using \texttt{qwen-vl-utils}.

For post-training with AI feedback, we use \texttt{gpt-5-mini-2025-08-07} with the reasoning level set to \texttt{low}. This setting was chosen to improve output consistency and reduce the effect of variance across repeated evaluations.

For the inference efficiency and deployment performance experiments, we use an NVIDIA Jetson AGX Orin running JetPack 6.2.1 as the base environment. For inference, we use bf16 precision and PyTorch scaled dot-product attention (SDPA), without FlashAttention-2 or additional deployment-specific optimizations such as quantization or TensorRT conversion. At inference time, each sample is processed using a video and a text query as input. We break the \textit{runtime} into several stages, including prompt construction, visual input processing, input encoding, model generation, and output decoding. We record both the runtime of each stage and the total runtime for each sample. 

\subsection{LLM-based Evaluation and TAU-GRPO Reward Function} 
\begin{figure}[!t]
  \centering
  \includegraphics[width=\textwidth]{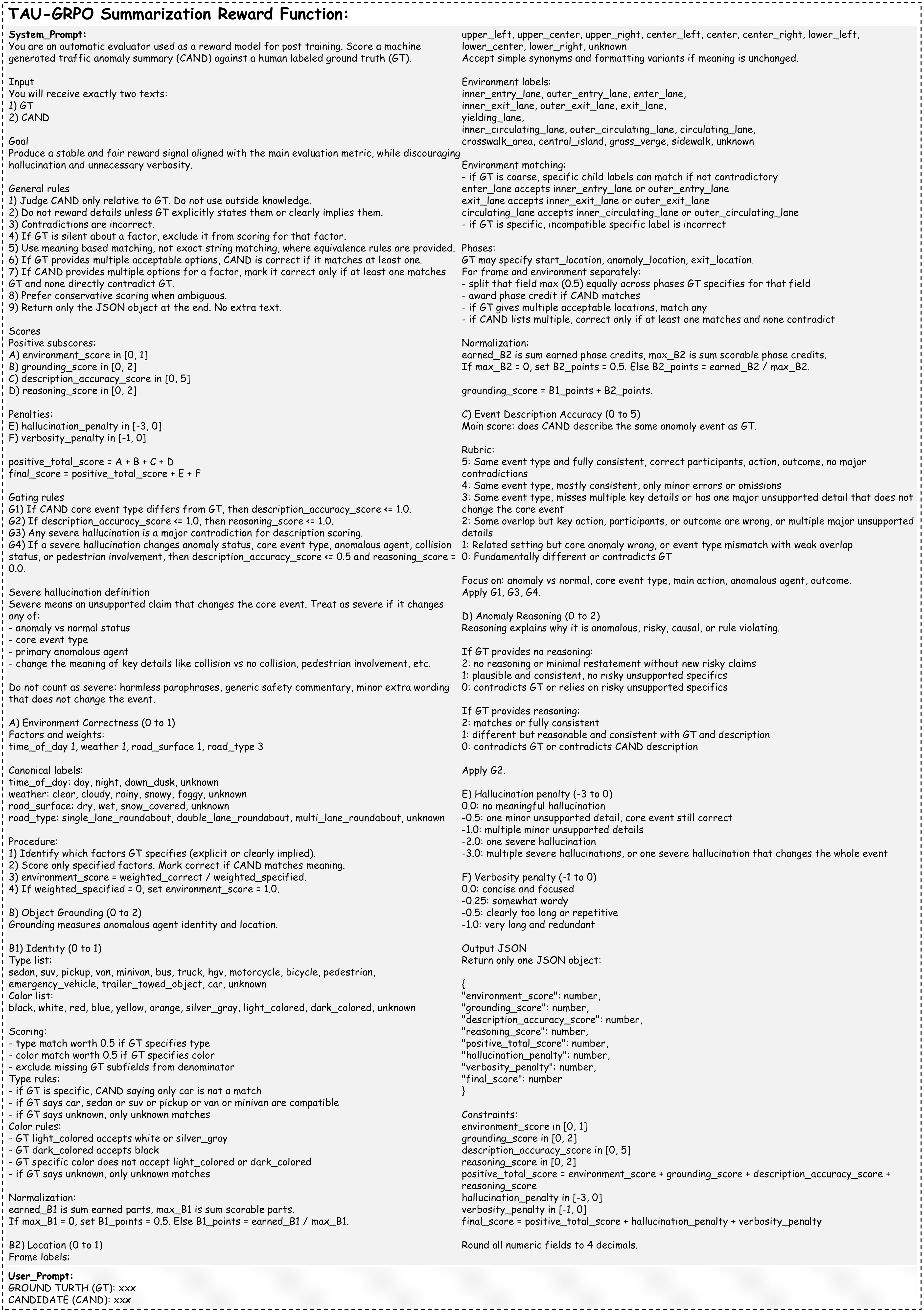}
  \caption{Prompt template for the TAU-GRPO reward function.}
  \label{fig:grpo-reward-function}
\end{figure}

In this section, we provide the full prompts used for LLM-based evaluation and the TAU-GRPO reward function in the anomaly summarization task. The LLM-based evaluation prompt is shown in \cref{fig:gpt-based eval}, while the TAU-GRPO reward prompt is shown in \cref{fig:grpo-reward-function}.

\subsubsection{LLM-based Evaluation}
The goal of LLM-based evaluation is to quantify the quality of the anomaly summary generated by the summarizer. Therefore, we evaluate each candidate summary from four aspects: (1) environment correctness, (2) object grounding, (3) event description accuracy, and (4) anomaly reasoning. Among these four aspects, environment correctness and object grounding can be defined more explicitly and scored in a structured manner. We formulate them as follows.

\textit{Environment correctness.}
Let $\mathcal{F}=\{\text{time},\text{weather},\text{surface},\text{road}\}$ denote the set of environment factors, with weights $w_{\text{time}}=1$, $w_{\text{weather}}=1$, $w_{\text{surface}}=1$, and $w_{\text{road}}=3$. For each factor $f\in\mathcal{F}$, let $s_f\in\{0,1\}$ indicate whether that factor is specified in the ground truth (GT), and let $c_f\in\{0,1\}$ indicate whether the candidate summary (CAND) matches GT on that factor. The environment correctness score is defined as
\[
\mathrm{EnvScore}=
\begin{cases}
1, & \sum\limits_{f\in\mathcal{F}} w_f s_f = 0, \\[6pt]
\dfrac{\sum\limits_{f\in\mathcal{F}} w_f s_f c_f}{\sum\limits_{f\in\mathcal{F}} w_f s_f}, & \text{otherwise}.
\end{cases}
\]

\textit{Object grounding.}
The object grounding score consists of two components, namely anomalous-agent identity grounding and anomalous-agent location grounding, $\mathrm{GroundingScore}=B_1+B_2$, where $B_1$ denotes identity grounding and $B_2$ denotes location grounding.

For identity grounding, the evaluation measures whether the candidate summary correctly identifies the anomalous agent's \emph{type} and \emph{color}. Let $\mathcal{I}=\{\text{type},\text{color}\}$ denote the set of identity attributes. For each attribute $a\in\mathcal{I}$, let $s_a\in\{0,1\}$ indicate whether that attribute is specified in GT, and let $c_a\in\{0,1\}$ indicate whether CAND matches GT on that attribute under the predefined compatibility rules. Since each attribute contributes equally, the identity grounding score is defined as
\[
B_1=
\begin{cases}
0.5, & \sum\limits_{a\in\mathcal{I}} s_a = 0, \\[6pt]
\dfrac{\sum\limits_{a\in\mathcal{I}} s_a c_a}
{\sum\limits_{a\in\mathcal{I}} s_a}, & \text{otherwise}.
\end{cases}
\]

For location grounding, the evaluation measures whether the candidate summary correctly identifies the anomalous agent's position in both the image frame and the road environment. Let $\mathcal{L}=\{\text{frame},\text{environment}\}$ denote the set of location fields. For each field $\ell\in\mathcal{L}$, let $s_\ell\in\{0,1\}$ indicate whether that field is specified in GT, and let $c_\ell\in\{0,1\}$ indicate whether CAND matches GT on that field under the predefined compatibility rules. Since each location field contributes equally, the location grounding score is defined as
\[
B_2=
\begin{cases}
0.5, & \sum\limits_{\ell\in\mathcal{L}} s_\ell = 0, \\[6pt]
\dfrac{\sum\limits_{\ell\in\mathcal{L}} s_\ell c_\ell}
{\sum\limits_{\ell\in\mathcal{L}} s_\ell}, & \text{otherwise}.
\end{cases}
\]
If multiple annotated phases are provided for a location field, the corresponding match variable is determined by aggregating phase-level matches according to the prompt rules.

For the remaining two aspects, \textit{event description accuracy} and \textit{anomaly reasoning}, it is difficult to design a reliable deterministic metric because both involve open-ended semantic alignment. Therefore, instead of defining them with explicit equations, we instruct the LLM to make a careful judgment based on consistency with GT, event-level correctness, and the absence of unsupported details.

\subsubsection{TAU-GRPO Summarization Reward Function}
The TAU-GRPO reward function follows the overall design of the LLM-based evaluation prompt, using the same four aspects as the basis for positive rewards. However, compared with LLM-based evaluation, the reward function is intentionally designed to be more conservative and stricter for event description accuracy and anomaly reasoning. In addition, two extra penalty terms, namely \textit{hallucination penalty} and \textit{verbosity penalty}, are introduced into the final reward. The main motivation for this stricter design is that reinforcement learning requires a sharper reward signal than post-hoc evaluation. In particular, we seek to minimize reward leakage to hallucinated, weakly aligned, or partially incorrect summaries, while encouraging concise and faithful outputs. By contrast, the LLM-based evaluation metric is intended to measure summary quality more comprehensively and therefore still assigns partial credit to partially correct predictions.

\begin{figure}[!t]
  \centering
  \includegraphics[width=\textwidth]{fig/llm-judge.png}
  \caption{Prompt template for LLM-based evaluation.}
  \label{fig:gpt-based eval}
\end{figure}

\subsection{Additional Qualitative Results}
\begin{figure}[h]
  \centering
  \includegraphics[width=\textwidth]{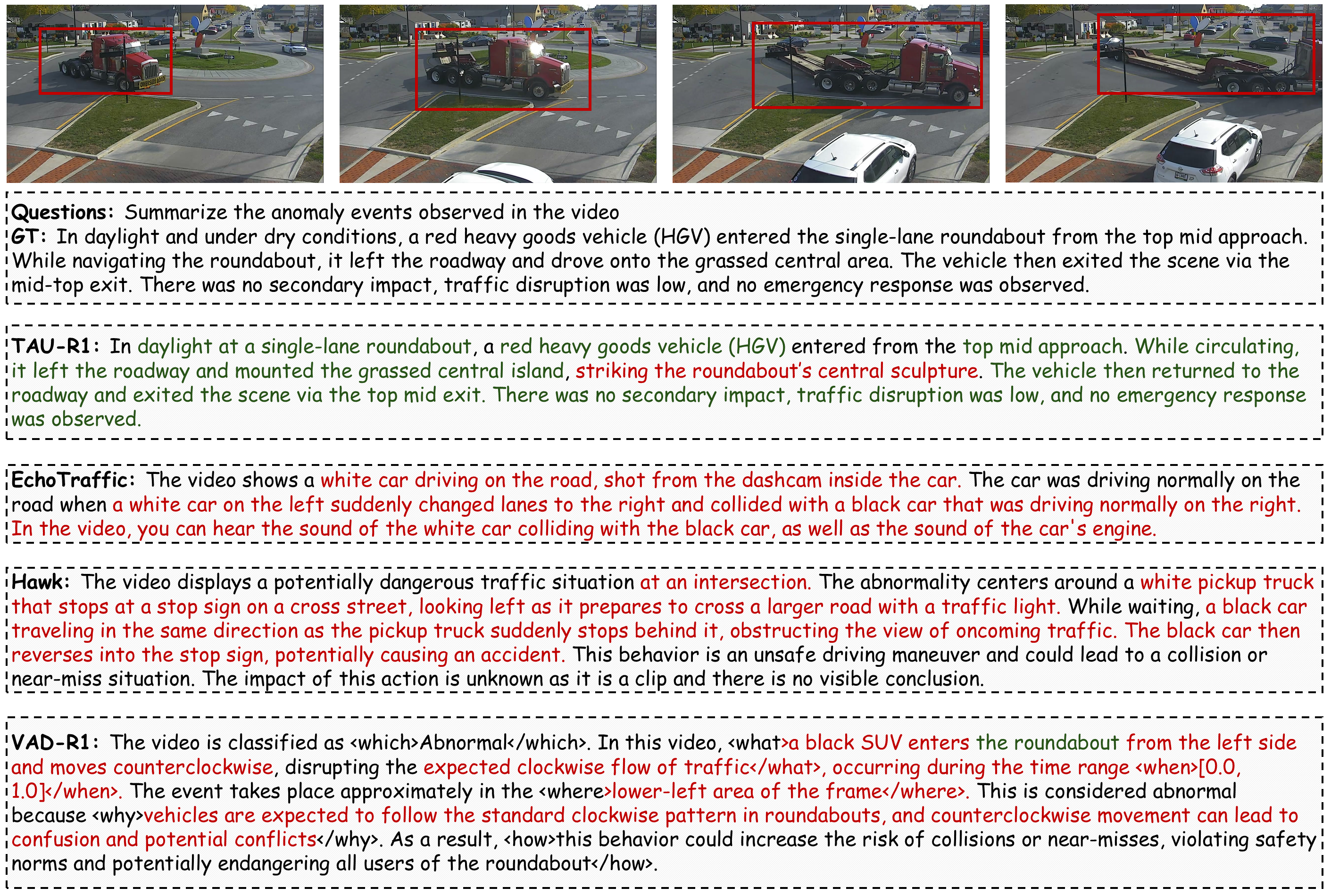}
  \caption{Additional qualitative example where TAU-R1 successfully identifies the anomaly in a long video, while competing VAU models fail to capture the key event.}
  \label{fig:more-qualitative1}
\end{figure}

\begin{figure}[h]
  \centering
  \includegraphics[width=\textwidth]{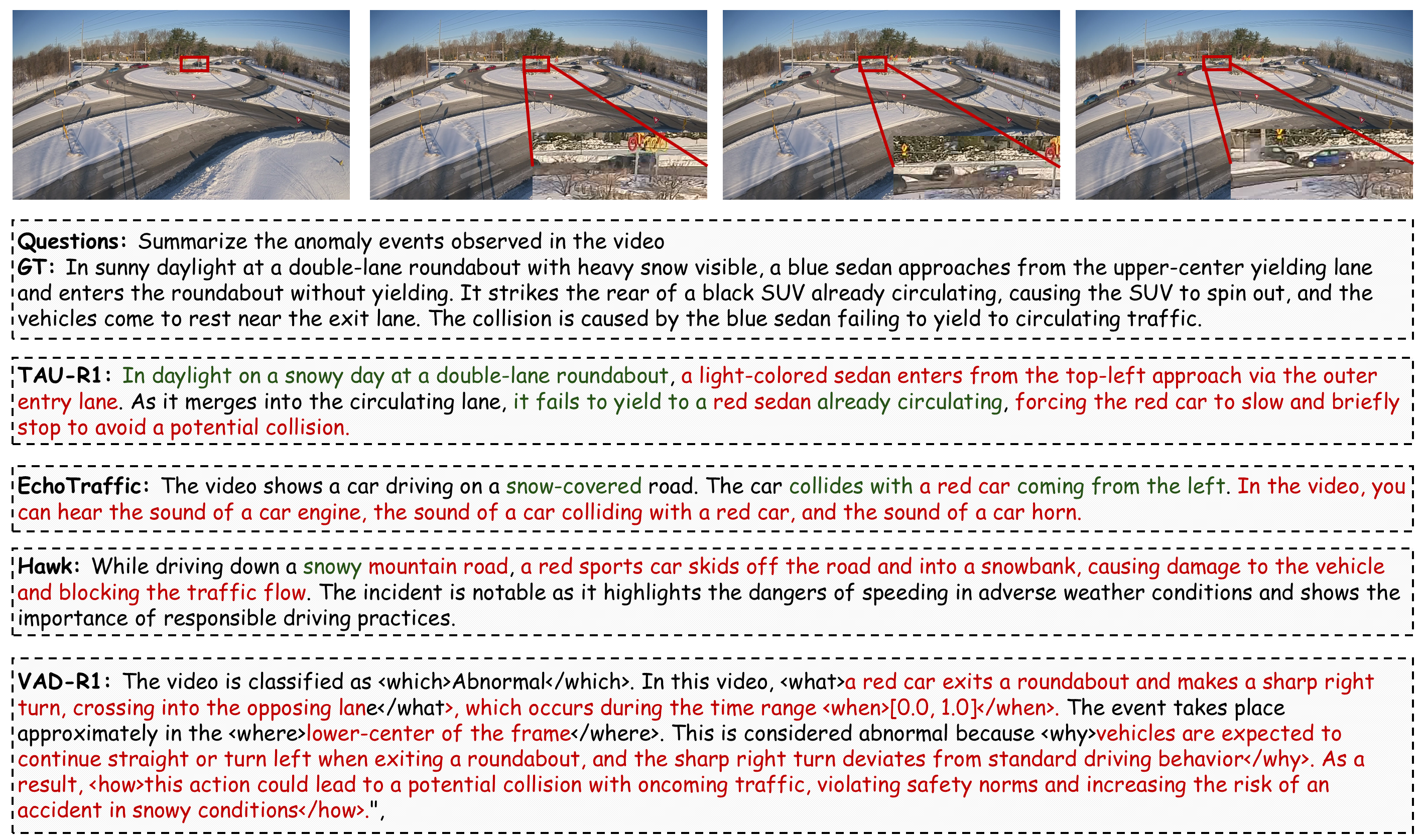}
  \caption{Additional qualitative example in a challenging far-field and partially occluded scene, where all models fail to fully capture the anomaly event.}
  \label{fig:more-qualitative2}
\end{figure}

We present two additional cases in \cref{fig:more-qualitative1} and \cref{fig:more-qualitative2} to further illustrate both the strengths and limitations of TAU-R1 under different levels of scene difficulty. The first example, shown in \cref{fig:more-qualitative1}, is a long video clip (210 seconds) containing a relatively brief anomalous event (32 seconds). In this example, TAU-R1 successfully identifies and summarizes the anomalous event, while the other VAU models fail to capture the key anomaly. The second example, shown in \cref{fig:more-qualitative2}, is substantially more challenging. The anomaly occurs in the far field of the camera view, making the anomalous objects very small. In addition, the event is partially affected by occlusion due to the camera viewing angle. Under this challenging setting, TAU-R1 fails to fully identify the key anomaly event, and the competing models also do not capture it correctly. 
    
\subsection{Qualitative Evaluation on Out-of-Domain Traffic Scenes}
\begin{figure}[t]
  \centering
  \includegraphics[width=\textwidth]{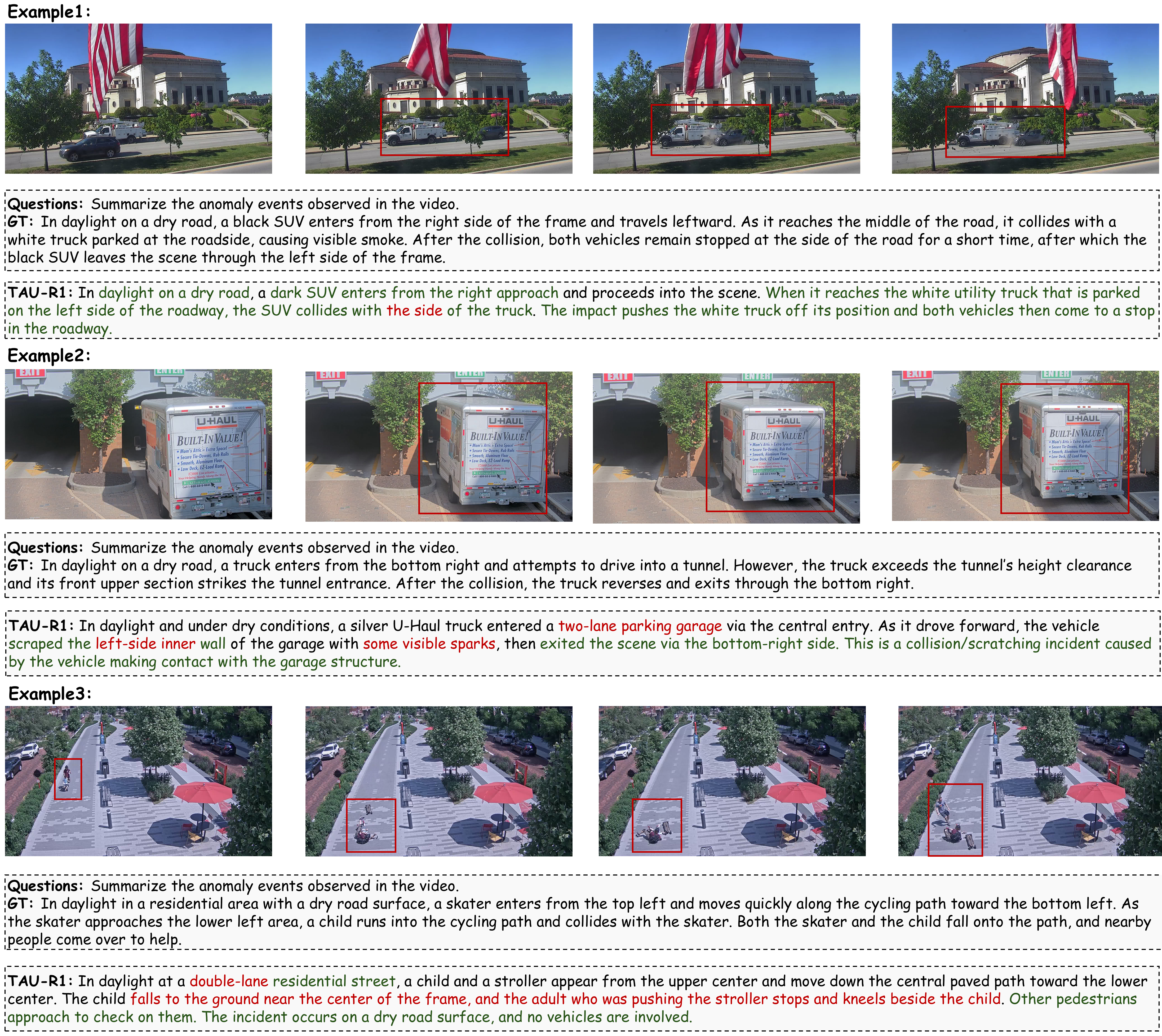}
  \caption{Out-of-domain traffic scene qualitative examples for TAU-R1.}
  \label{fig:zero-shot-qualititive}
\end{figure}

We provide three additional qualitative examples in \cref{fig:zero-shot-qualititive}, using traffic anomaly videos captured in the City of Carmel outside the roundabout scenes included in Roundabout-TAU. All three examples were excluded from model training. Example 1 corresponds to a roadside camera capturing a collision event with limited anomaly candidates. In this example, TAU-R1 correctly identifies and summarizes the event, demonstrating its ability to summarize out-of-domain traffic anomalies. Example 2 is a more challenging case, involving a less common environment (a garage tunnel) and a relatively subtle anomaly (a truck striking the tunnel entrance). Here, the model captures the main event correctly, but also introduces several hallucinated details, indicating reduced reliability under more ambiguous conditions. Example 3 is even more challenging: the anomaly occurs in a residential-area-like scene rather than on the main roadway, and the primary agents are pedestrians or cyclists rather than vehicles. Despite this substantial domain shift, TAU-R1 still captures part of the key event semantics. Overall, these zero-shot examples suggest that TAU-R1 learns transferable traffic-scene reasoning patterns rather than overfitting to the roundabout scenarios seen during training. 

\section{Limitations and Future Work}
In this paper, we introduce the Roundabout-TAU dataset and the TAU-R1 framework to address traffic anomaly understanding. We also evaluate its deployment efficiency by deploying the model on an NVIDIA Jetson edge device, highlighting its potential for future real-world ITS applications.

Despite its strong performance over existing baselines, TAU-R1 still has several limitations. First, the model may struggle with complex and challenging traffic events, especially when the anomalous objects are far from the roadside cameras and therefore appear very small. Second, although Roundabout-TAU provides a realistic benchmark for roadside traffic anomaly understanding, it is still focused on roundabout-centered scenes, which may limit generalization to other traffic environments with substantially different layouts, object compositions, or activity patterns.

In future work, we plan to improve the robustness of TAU-R1 for more complex anomaly scenarios and broaden its generalization ability across diverse traffic scenes. We also aim to move beyond the TAU task by integrating TAU-R1 with other ITS modules, such as multi-camera tracking, traffic monitoring, and decision-support systems, to support safer and smarter real-world intelligent transportation systems.

\newpage
\end{document}